\documentclass[10pt,twocolumn,letterpaper]{article}

\usepackage[pagenumbers]{cvpr} 

\usepackage{amsmath,amsfonts,bm}

\def\eqref#1{equation~\ref{#1}}

\def\1{\bm{1}}

\DeclareMathAlphabet{\mathsfit}{\encodingdefault}{\sfdefault}{m}{sl}
\SetMathAlphabet{\mathsfit}{bold}{\encodingdefault}{\sfdefault}{bx}{n}

\DeclareMathOperator*{\argmin}{arg\,min}

\usepackage{url}
\usepackage[T1]{fontenc}    
\usepackage{url}            
\usepackage{booktabs}       
\usepackage{amsfonts}       
\usepackage{nicefrac}       
\usepackage{microtype}      
\usepackage{multirow}
\usepackage{tikz}
\usetikzlibrary{calc}
\usepackage{comment}
\usepackage{caption}
\usepackage{subcaption}
\usepackage{graphicx}
\usepackage{soul}
\usepackage{amsmath, amsthm,amssymb}
\usepackage{pifont}
\usepackage{inconsolata}
\usepackage{tabto}
\usepackage[linesnumbered,ruled,vlined]{algorithm2e}
\usepackage {algpseudocode}
\usepackage{algorithmicx}
\usepackage{algcompatible}
\usepackage{xfrac}   
\usepackage{adjustbox}
\usepackage{wrapfig}
\usepackage[normalem]{ulem}
\usepackage{pgfplots}
\pgfplotsset{compat=1.7}
\usepackage{enumitem}
\usepackage[most]{tcolorbox}

\usepackage{lipsum}

\definecolor{cvprblue}{rgb}{0.21,0.49,0.74}
\usepackage[pagebackref,breaklinks,colorlinks,allcolors=cvprblue]{hyperref}

\def\paperID{23} 
\def\confName{CVPR}
\def\confYear{2025}

\title{Dealing with the Evil Twins: Improving Random Augmentation by Addressing Catastrophic Forgetting of Diverse Augmentations}

\author{Dongkyu Cho\\
New York University\\
New York, New York\\
{\tt\small dongkyu.cho@nyu.edu}
\and
Rumi Chunara\\
New York University\\
New York, New York\\
{\tt\small rumi.chunara@nyu.edu}
}

\begin{document}
\maketitle

\begin{abstract}

Data augmentation is a promising tool for enhancing out-of-distribution generalization, where the key is to produce diverse, challenging variations of the source domain via costly targeted augmentations that maximize its generalization effect. Conversely, random augmentation is inexpensive but is deemed suboptimal due to its limited effect. In this paper, we revisit random augmentation and explore methods to address its shortcomings. We show that the stochastic nature of random augmentation can produce a set of colliding augmentations that distorts the learned features, similar to catastrophic forgetting \citep{Kirkpatrick_2017}. We propose a simple solution that improves the generalization effect of random augmentation by addressing forgetting, which displays strong generalization performance across various single source domain generalization (sDG) benchmarks.

\end{abstract}

\section{Introduction}
\label{sec:intro}
\begin{figure} 
    \centering
    \vspace{-3mm}
    \includegraphics[width=0.49\textwidth]{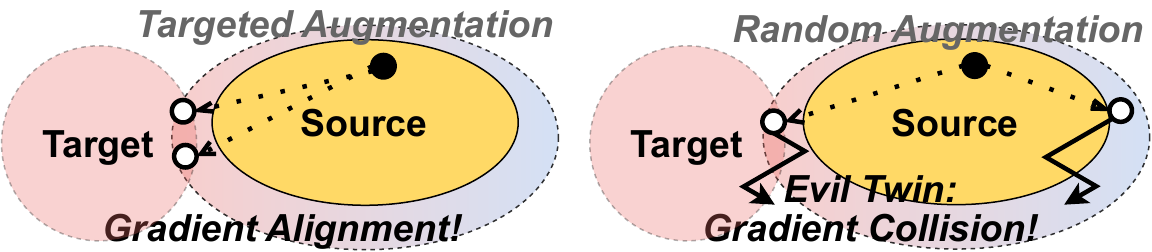}
    \caption{Data Augmentation enhances generalization by generating samples closer to the target data. However, contradicting augmentations can induce gradient collisions, leading to forgetting.}
    \label{fig:sdg}  
    \vspace{-5mm}
\end{figure}

Learning a robust model from a single source domain mirrors many real-world scenarios where a machine learning model has limited access to training data but is expected to perform well across various out-of-distribution (OOD) test data \citep{efthymiadis2024craftingdistributionshiftsvalidation}. Previous studies have shown that expanding the source domain through data augmentation \citep{qiao2020} improves generalization on unseen target domains. \cref{fig:sdg} illustrates the general concept of these augmentation-based generalization methods. Many of these methods rely on targeted augmentation techniques that select transformations based on designated objectives (e.g., adversarial loss \citep{volpi2018, zheng2024advst}), owing to its efficacy in improving the model's OOD generalization capabilities. In contrast, random augmentation selects transformations stochastically, while showing a limited generalization effect \citep{aminbeidokhti2024domain}. 


In this paper, we answer the following question: "Why does random augmentation underperform?" despite its ability to efficiently provide challenging, diverse augmentations. An important motivation for this work is that cost-effective data augmentation is crucial, especially in the era of large foundation models, where training and inference resources are already at a premium. Targeted augmentation methods, while effective, generally demand additional inference passes through the main model as well as an auxiliary module to optimize or select transformations, incurring significant computational overhead. In contrast, random augmentation imposes minimal extra cost during training yet still ensures a broad range of transformations.

We hypothesize that the underperformance of random augmentation stems from its stochastic nature, where colliding augmentations of the same source-- evil twin--distort the model's trained features \citep{Kirkpatrick_2017}. Throughout this paper, we theoretically demonstrate that the diversity of the random augmentation stimulates gradient collision \citep{saha2023continual}, distorting learned features (i.e., forgetting). We substantiate our hypothesis through empirical analysis and show that conventional continual learning (CL) algorithms can mitigate this phenomenon, but are generally unsuitable. Reflecting on this, we propose a simple solution that directly manipulates the model's weight space to accumulate the learned features of colliding augmentations. 

Our contributions are summarized as follows:
\begin{itemize}
    \item We theoretically show that diverse augmentations can induce a unique variant of catastrophic forgetting induced by colliding augmentations (i.e., evil twin).
    \item We show that the generalization effect of random augmentation can be enhanced by addressing forgetting, namely through continual learning algorithms.
    \item We present a simple method that substitutes replay memory using model weights.
\end{itemize}

\section{Related Works}
\label{sec:preliminaries}
\paragraph{Data Augmentation: Targeted vs. Random}

Data augmentation is a common technique for improving model robustness by improving data diversity \citep{geiping2022much}. It is instrumental when the training data is limited or lacks diversity \citep{he2019dataaugmentationrevisitedrethinking,balestriero2022data}, such as in cases of single-source domain generalization \citep{efthymiadis2024craftingdistributionshiftsvalidation}. At large, two categories of data augmentation techniques are used: targeted augmentation and random augmentation. Targeted augmentation techniques aim to generate challenging augmentations, namely via providing adversarial examples \citep{goodfellow2015explainingharnessingadversarialexamples,xie2020adversarialexamplesimproveimage}. However, this family of data augmentation is computationally demanding due to the additional cost required for adversarial training. On the other hand, random augmentation applies a random combination of transformations (e.g., flipping, noise injection, etc.) \citep{cubuk2020randaugment,cubuk2018autoaugment,efthymiadis2024craftingdistributionshiftsvalidation}. Compared to targeted augmentation techniques, random augmentation does not require additional optimization processes but shows limited generalization effect \citep{aminbeidokhti2024domain}.

\paragraph{Single Source Domain Generalization}
In single-source domain generalization (sDG), only one domain is available for training, limiting the ability to exploit domain differences for robustness. To address this, many methods simulate diverse domains via data augmentation. For example,  \citet{volpi2018} adopted adversarial learning strategies to generate complex, label-preserving perturbations, and \citet{zhao2020} introduced entropy-based regularization to generate challenging augmentations. These strategies inspired subsequent approaches \citep{li2021,wang2021,xu2023simde,zheng2024advst}. Meanwhile, \citet{efthymiadis2024craftingdistributionshiftsvalidation} introduced a novel validation strategy for sDG. In this work, we evaluate the efficacy of data augmentation on standard sDG benchmarks.

\paragraph{Catastrophic Forgetting \& Continual Learning}
Catastrophic forgetting \citep{goodfellow2013empirical} is a common phenomenon in machine learning, particularly in the context of continual learning \citep{Kirkpatrick_2017,wang2302comprehensive}. Forgetting occurs when a pre-trained model forgets previously learned features when learning a new domain. While most studies focus on catastrophic forgetting in a multi-domain setting \cite{rolnick2019experience, chaudhry2019tiny, wang2022foster}, our work addresses a lesser-explored issue: catastrophic forgetting exacerbated by data augmentation, in a single domain setting \citep{guo2023out, ma2022continual, rostami2019complementary,cho2025peer}.

\section{Evil Twin Effect: Random Augmentation Exacerbates Catastrophic Forgetting of diverse augmentations.}
\label{sec:limitation}

In this section, we analyze the generalization effect of random augmentation and reveal its overlooked issues. In the sequel, we show that (1) random augmentation provides diverse transformations that benefit generalization, but (2) the diversity of the random augmentation produces contradicting augmentations that exacerbate gradient collision (i.e., evil twin), fostering catastrophic forgetting (3) lastly, we show that conventional continual learning algorithms \textit{generally} cannot address the issue.

\paragraph{Notation and Setup}
We begin with the notations. Let $x \sim D$ denote the original sample (e.g., image) sampled from the dataset $D$, and let the finite transformation set $T= \{T_1, T_2, \cdots T_n \}$, represent the set of all $n$ available augmentation transformations that can be applied to $x$. For simplicity, we assume there is a finite set of transformations that can be applied to the original sample, which reflects the reality of many existing augmentation algorithms \citep{cubuk2020randaugment,efthymiadis2024craftingdistributionshiftsvalidation}

At the $i$th step of the training stage, random augmentation $\tau_{ra}$ selects a transformation $T_i \in T$ with a uniform probability, expressed as: 
\begin{equation}\label{eq:rand_uniform}
    P_{\tau_{ra}}(T_i) = \frac{1}{n} , \forall T_i \in T.
\end{equation}    

The targeted augmentation $\tau_{l}$ selects a transformation $T_{l}$ that minimizes a predefined objective $\mathcal{L}(\theta;T_i(x))$ -- namely adversarial loss \citep{volpi2018} -- over the model parameters $\theta$: 
\begin{equation}\label{eq:target_nonuniform}
    P_{\tau_{l}}(T_i) = \frac{e^{-\beta\mathcal{L}(\theta;T_i(x))}}{\sum\nolimits_{j=1}^{n} e^{-\beta\mathcal{L}(\theta;T_j(x))}},
\end{equation}   

where $\beta > 0 $ is a hyperparameter that adjusts how sensitive the probabilities are influenced by the loss $\mathcal{L}$. For instance, setting $\beta= 0$ would make $\tau_{l}$ equivalent to $\tau_{ra}$, as all the transformations would have equal probability. Due to this optimization, transformations under $\tau_{l}$ are selected based on alignment with $\mathcal{L}$, resulting in a non-uniform probability distribution.

\subsection{Quantifying Augmentation Diversity: Random vs. Targeted Augmentation}\label{sec:3-1}

In this section, we show that random augmentation provides a more diverse set of augmentations compared to the targeted augmentation strategies. 

\paragraph{Comparing Augmentation Diversity}

To measure the diversity of the augmentation's outputs (i.e., augmentation diversity), we use the entropy $H$ of the transformation distribution induced by $\tau_{ra}$ and $\tau_{l}$. The entropy $H(P)$ over a discrete probability distribution $P$ over transformations $T$ is defined as: 
\begin{equation}
    H(P) = - \sum_{i=1}^{n} P(T_i)\log P(T_i),
\end{equation}    
where higher entropy values indicate greater stochasticity and thus, greater diversity in the transformations applied. By comparing the entropy of the two augmentation strategies, we can show that the entropy of targeted augmentation is lower than that of random augmentation. 

\paragraph{Proof Sketch.}
Recall that \(\tau_{ra}\) selects each transformation \(T_i \in \{T_1,\dots,T_n\}\) with uniform probability \cref{eq:rand_uniform}. Consequently, its entropy satisfies
\begin{equation}
H(\tau_{ra}) \;=\; 
- \sum_{i=1}^n \frac{1}{n} \log\Bigl(\tfrac{1}{n}\Bigr) 
\;=\; \log n.    
\end{equation}

In contrast, the targeted augmentation \(\tau_{l}\) assigns probabilities based on the loss \cref{eq:target_nonuniform},
leading to a \emph{non-uniform} distribution and thus lower entropy. Consequently,
\begin{equation}
H(\tau_{l}) \;<\; \log n \;=\; H(\tau_{ra}),
\end{equation}
which establishes that random augmentation \(\tau_{ra}\) indeed induces higher entropy, and thus more diverse transformations, than the targeted augmentation \(\tau_{l}\). For the full derivation, refer to \cref{appendix:augmentation_diversity}.

\begin{tcolorbox}[colframe=black, colback=blue!2!, coltitle=black, width=\linewidth, boxrule=0.5mm]
    \textbf{Takeaway:} Random Augmentation introduces greater augmentation diversity than any targeted augmentation that relies on a loss objective.
\end{tcolorbox}

\subsection{The Impact of Augmentation Diversity on Catastrophic Forgetting}\label{sec:3-2}

Next, we analyze how higher diversity from random augmentation can lead to increased catastrophic forgetting \citep{Kirkpatrick_2017}, compared to targeted augmentation methods with lower augmentation diversity. Specifically, we show that a high augmentation diversity fosters gradient collisions that erode the model's trained features during parameter updates.

\paragraph{Proof Sketch} 
Let \(g_x = \nabla_\theta L(\theta; x)\) be the gradient computed on the original sample \(x\) and \(g_{T(x)} = \nabla_\theta L(\theta; T(x))\) be the gradient computed on the augmented sample \(T(x)\). Assume that the augmentation introduces a perturbation \(\delta\) such that
\begin{equation}
T(x) = x + \delta,
\end{equation}
with \(\delta\) drawn from a distribution whose variance reflects the diversity of the augmentation. Using a first-order Taylor expansion, we approximate
\begin{equation}
g_{T(x)} \approx g_x + M\,\delta,
\end{equation}
where \(M = \nabla_\theta \nabla_x L(\theta;x)\) is the matrix of mixed second-order derivatives.

We then analyze the alignment of the gradients via their cosine similarity:
\begin{equation}
\cos(\phi) = \frac{g_x^\top (g_x + M\,\delta)}{\|g_x\|\cdot\|g_x + M\,\delta\|}.
\end{equation}
For random augmentation, the perturbation \(\delta\) has high variance, which increases the magnitude of the term \(M\,\delta\) and hence introduces greater fluctuations in \(g_{T(x)}\). Consequently, when taking the expectation over \(\delta\), the expected cosine similarity \(\mathbb{E}[\cos(\phi)]\) is lower, signifying that the gradients are less aligned (i.e., more likely to collide) compared to the case of targeted augmentation, where \(\delta\) is smaller.

This misalignment—referred to as \emph{gradient collision}—leads to destructive interference during parameter updates, which gradually erodes the model's learned features \citep{yu2020gradient,chen2020just,cho2025peer}. For the full derivation, please see \cref{appendix:gradient_collision}.

\begin{tcolorbox}[colframe=black, colback=blue!2!, coltitle=black, width=\linewidth, boxrule=0.5mm]
    \textbf{Takeaway:} The augmentation diversity of Random Augmentation exacerbates forgetting by triggering gradient collision.
\end{tcolorbox}

\subsection{Empirical Analysis of the Problem and Solutions}

In this section, we provide an empirical analysis of the problem and seek practical ways to address it for the improvement of random augmentation. First, we empirically show the effect of the forgetting syndrome on the model's generalization. Specifically, we study (1) its effect on the prediction of the model, and (2) the model's learned features.

\begin{figure}
    \centering
    \vspace{-5mm}
    \includegraphics[clip, trim=1.6cm 3mm 1.6cm 1cm, height=3cm]{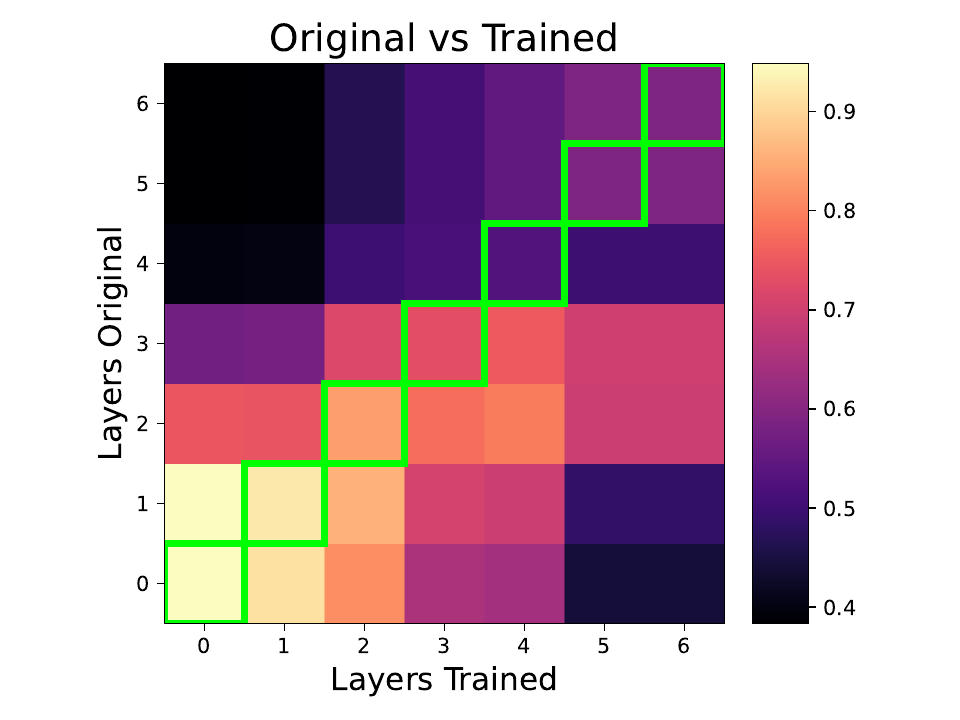}
    \caption{CKA Similarity between models trained with two distinct augmentations of the same dataset.}\label{fig:cka_augmentation}
\end{figure}

To assess the effect of the colliding augmentations on the catastrophic forgetting, we design an experiment that uses a simple transformation (e.g., rotation) to simulate "evil twin", a set of augmented samples that results in colliding gradients in the parameter space \citep{yu2020gradient}, and study the model's behavior when it is trained on such pairs. Specifically, we test a model's prediction accuracy before and after it is trained on a set of evil twin. 

\begin{figure*}[t!]
\centering
\begin{minipage}[t]{0.44\textwidth}
\centering
\captionof{table}{Effect of evil twin on catastrophic forgetting. Model sequentially trained on colliding augmentations (i.e., evil twin) suffer from significant forgetting of the previous augmentations, which limit augmentation's generalization effect.
} 
\fontsize{9}{10}\selectfont
\centering
\label{tab:evil_twins}
\begin{adjustbox}{width=0.66\textwidth}
\begin{tabular}{@{}lccc@{}}
\toprule
{Rotation} & $\overline{\mathrm{SD}}$ & $x_1$ Acc & Forgetting \\ 
\midrule
0$^\circ$  & 0.261 & 0.987 & 0.0013 \\
-10$^\circ$  & 0.272 & 0.985 & 0.0031 \\
-20$^\circ$  & 0.290 & 0.965 & 0.0236 \\
-30$^\circ$  & 0.306 & 0.921 & 0.0668 \\
-40$^\circ$  & 0.315 & 0.797 & 0.1916 \\
-50$^\circ$  & 0.330 & 0.624 & 0.3638 \\
-60$^\circ$ & 0.337 & 0.532 & 0.4557 \\
-70$^\circ$ & 0.340 & 0.389 & 0.5989 \\
-80$^\circ$ & 0.340 & 0.308 & 0.6803 \\
-90$^\circ$ & 0.341 & 0.252 & 0.7365 \\
\bottomrule
\end{tabular}

\end{adjustbox}
\end{minipage}
\hfil
\begin{minipage}[t]{0.47\textwidth}
\centering
\vspace{0pt}
\includegraphics[width=0.6\textwidth]{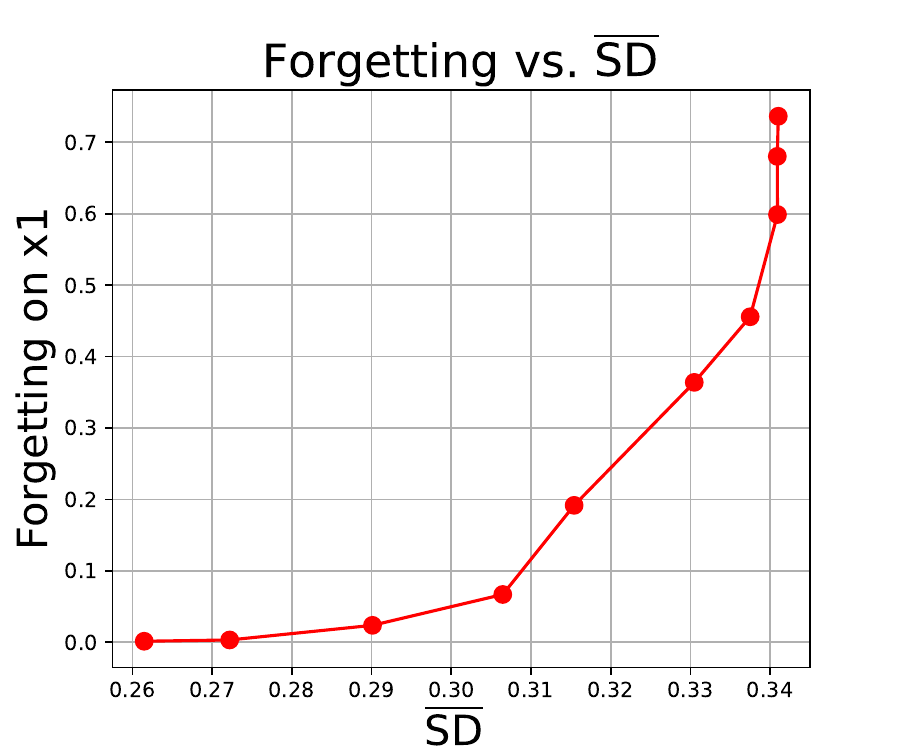}
\caption{Aggregated Sign Discrepancy ($\overline{\mathrm{SD}}$) vs. Forgetting. Increase in the level of collision between augmented samples (measured as $\overline{\mathrm{SD}}$) coincides with the worsening of the forgetting.} \label{fig:evil_twins_forgetting}

\end{minipage}
\end{figure*}

\paragraph{A formal definition of Evil Twin}
Let $\nabla f_1, \nabla f_2 \in \mathbb{R}^d$ be the gradients of the loss function with respect to the model parameters computed from different augmentations of the same data sample (or batch). In our notation, we write
\begin{equation}
\mathbf{g}_1 = \nabla f_1 \quad \text{and} \quad \mathbf{g}_2 = \nabla f_2. 
\end{equation}

We say that $(\mathbf{g}_1,\mathbf{g}_2)$ are \emph{evil twin} if their parameter-wise signs differ extensively, thereby inducing catastrophic interference in the model's parameter updates.

Formally, we define the \emph{sign discrepancy} between $\mathbf{g}_1$ and $\mathbf{g}_2$ as
\begin{equation}
\label{eq:sign_discrepancy}
  \mathrm{SD}(\mathbf{g}_1, \mathbf{g}_2) \;=\; 
  \frac{1}{d}\sum_{i=1}^{d} \mathbf{1}\Bigl(\operatorname{sign}(g_{1,i}) \neq \operatorname{sign}(g_{2,i})\Bigr),
\end{equation}
where $g_{1,i}$ denotes the $i$th component of $\mathbf{g}_1$, and $d$ is the total number of parameters.

In practice, to reduce noise from a single mini-batch, we aggregate the sign discrepancy over $k$ mini-batches. Let $\{(\mathbf{x}_b, \mathbf{y}_b)\}_{b=1}^k$ be $k$ mini-batches sampled from the same augmentation, and let $\mathbf{g}_{2,b}$ denote the gradient computed on batch $b$ using the second augmentation. We then define the aggregated sign discrepancy as
\begin{equation}
\label{eq:aggregated_sign_discrepancy}
  \overline{\mathrm{SD}}(\mathbf{g}_1, \{\mathbf{g}_{2,b}\}_{b=1}^k)
  \;=\;
  \frac{1}{k}\sum_{b=1}^k \mathrm{SD}\bigl(\mathbf{g}_1,\mathbf{g}_{2,b}\bigr).
\end{equation}

This aggregated measure $\overline{\mathrm{SD}}$ provides a smoother and more reliable indication of the level of parameter-wise sign conflict—hence, the extent of "Evil Twin" behavior—than that computed on a single mini-batch.

\paragraph{Effect on Prediction}

Using this measure (i.e., $\overline{\mathrm{SD}}$ ), we analyze the effect of the evil twin on the model's prediction in a formal fashion. Specifically, we demonstrate that two colliding augmentations $x_1,x_2$ of an original sample $x$ can distort the learned features of each other, causing them to forget what it has learned previously. To show this, we design a simple experiment using the MNIST dataset, where the model is first trained on a image rotated clock-wise $45^\circ$ ($x_1$), and then re-trained on another rotated image ($x_2$) with a different rotation value ranging from $45^\circ$ to $-45^\circ$ ($0^\circ$ to $-90^\circ$ compared to $x_1$). We then analyze the drop in $x_1$ accuracy (i.e., forgetting) and measure the aggregated sign discrepancy ($\overline{\mathrm{SD}}$), where $k=10$, meaning that we measure the $\overline{\mathrm{SD}}$ over 10 mini-batches. The results are reported in \Cref{tab:evil_twins} and \Cref{fig:evil_twins_forgetting}.

We can see a general trend in \Cref{tab:evil_twins}. As the aggregated sign discrepancy (i.e., evil twin metric) increases, the accuracy on the previously learned samples decreases (i.e., forgetting increases). While our analysis is based on rotation-based augmentations on MNIST and focuses on the effect of a single prior augmentation, the aggregated sign discrepancy metric is generalizable as is rooted in the fundamental dynamics of gradient-based optimization, quantifying parameter-level conflicts independent of augmentation type, and averaging over multiple mini-batches reduces noise. 

\paragraph{Effect on Learned Features}
Next, we analyze how the model's learned features are altered during training with random augmentation, namely by using the Centered Kernel Alignment (CKA) similarity metric \citep{kornblith2019similarity}. Specifically, we compare the intermediate output features of the model that is trained with augmented MNIST but with distinct, non-overlapping augmentations. In \cref{fig:cka_augmentation}, we visualize the results, where the diagonal cells of the box indicate the feature similarity between the corresponding layers between the two models, bright colors indicating high feature similarity. As we can see in \cref{fig:cka_augmentation}, the features learned from augmented data diminish as the model is exposed to distinct augmentations.

\paragraph{Solutions}
We then seek ways to address forgetting. Our initial idea is to view the evil twin effect as a continual learning (CL) problem. Naturally, we start by using an existing CL algorithm that explicitly mitigates forgetting.

Specifically, we implement a simple replay algorithm that saves subsets of augmented samples from previous steps to explicitly tackle forgetting via exemplar memory replay \citep{rolnick2019experience}. We denote this as the \textit{memory} method in our experiments. We find that the memory method effectively boosts performance while stabilizing the learning process (see \cref{fig:cka_erm_memory}). In \cref{sec:exp}, we have shared the details of the experimental results. However, this approach has some limitations. First, replay memory is only effective when the discrepancy between the augmented data and the original data is significant. In other words, when the gap between the augmented sample and the original sample is not significant, they are not reliable. Furthermore, to elicit stronger generalization, more than $50\%$ of the augmented samples were needed to be stored as the replay memory. In the sequel, we present a simple idea to address the forgetting problem, namely by substituting the memory samples with model weights.

\section{Method}
\label{sec:method}

\begin{algorithm}[t]
\caption{Our Method}
\label{alg:ours}
\footnotesize
\textbf{Input:} Task model $f$ with parameters $\theta$, source data $D_s$, snapshot interval $k$;  

\textbf{Output:} Fully updated task model $f$ and its parameters $\theta$

Save snapshot $\theta_s  \gets \theta$

\While{not converge}{
    \For{$i=1 : n_{iterations}$}{
        \If{$i \% k= 0$ }{
            Select layers to merge based on rank
            
            $\theta \gets \textsc{merge}(\theta, \theta_s)$ 
        
            Update snapshot $\theta_s \gets \theta$ 
            
            Randomly select augmentation $g$ 
                        
        }
        Sample $i$-th mini-batch from data $D_s$
        
        Augment the mini-batch using $g$
        
        Train $f$, update $\theta$
    }
}
\end{algorithm}


\begin{figure}
\vspace{-2mm}

    \centering
    \begin{subfigure}[b]{0.48\linewidth}
        \centering
        \includegraphics[height=5cm]{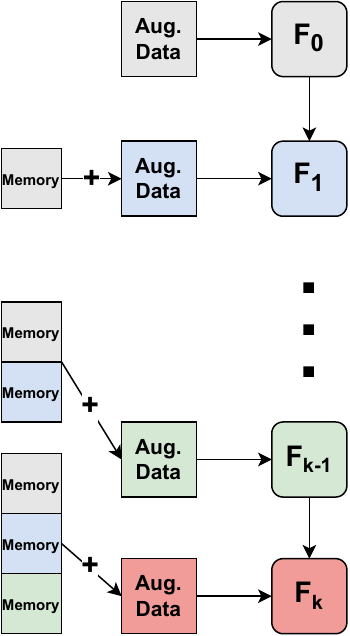}
        \caption{Memory Method}
    \end{subfigure}
    \hfill
    \begin{subfigure}[b]{0.48\linewidth}
        \centering
        \includegraphics[height=5cm]{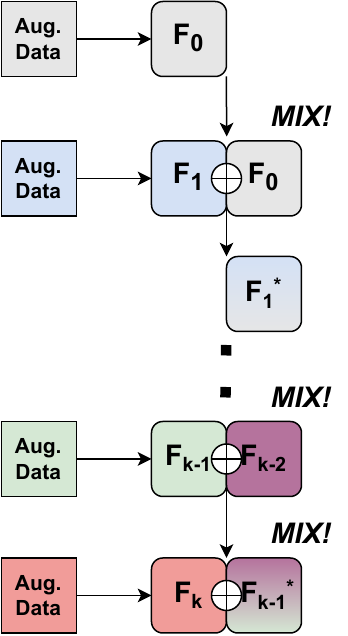}
        \caption{Ours}
    \end{subfigure}
    \caption{Replay memory helps address forgetting. In our scenario, we substitute replay memory with model weights.}
    \vspace{-5mm}
    \label{fig:framework}
\end{figure}

\paragraph{Notation and Setup} Our objective is to train a robust model $F= E\circ C$ using only the source domain dataset $D_{s}$. The model is composed of an encoder $E: \mathcal{X} \rightarrow \mathcal{H}$, and a classification head $C: \mathcal{H} \rightarrow \mathcal{Y}$. Additionally, we may implement a projection head $P: \mathcal{H} \rightarrow \mathcal{P}$. Specifically, the encoder receives the input image $x$ and returns a feature representation $h$, while the classification head predicts the prediction logits $\hat{y}$ from $h$. In this process, we will use an augmentation module $G$ that provides augmented samples $\bar{x}$ from the original sample $x$. Previous works that utilize learnable augmentation methods commonly utilize the predictions of the training model $F= E\circ C$ to update the augmentation module \citep{volpi2018}. On the other hand, our method utilizes random augmentation \citep{cubuk2020randaugment}, which alleviates the need for additional training costs.

\paragraph{Proposed Method} 

Let \(\theta \in \mathbb{R}^N\) be the current weights of the model and \(\theta_s \in \mathbb{R}^N\) be the snapshot of the weights saved \(k\) iterations earlier. Define the absolute difference for each parameter as
\begin{equation}\label{eq:drift}
d_i = \bigl| (\theta)_i - (\theta_s)_i \bigr|, \quad i=1,\dots,N.
\end{equation}
We then construct a binary mask \(\mathbf{m} \in \{0,1\}^N\) by selecting the indices corresponding to the top $p\%$ (in our case, $p=80$) of the differences:

\begin{equation}
m_i = \begin{cases}
1, & \text{if } d_i \geq \tau, \\
0, & \text{otherwise},
\end{cases}    
\end{equation}

where the threshold \(\tau\) is chosen so that
\begin{equation}
   \frac{1}{N}\sum_{i=1}^{N} m_i = \frac{p}{100}.
\end{equation}

The weight merging is then performed via a selective averaging of the current weights with the snapshot:
\begin{equation}
\theta \leftarrow \mathbf{m} \odot \frac{\theta + \theta_s}{2} \;+\; (1-\mathbf{m}) \odot \theta,    
\end{equation}

where $\odot$ denotes element-wise multiplication.

This expression updates only the \(80\%\) of the parameters that have experienced the largest shifts during training, effectively substituting replay memory with model weight averaging to counteract forgetting (the evil twin effect).

Our algorithm leverages the continual learning effects of model merging \citep{saha2023continual,marouf2023weighted}, to address forgetting. Specifically, we accomplish this by averaging the model's weights with the weight snapshot $\theta_s$ saved $k$ iterations prior. Our method utilizes $\theta_s$ as a substitute for replay memory, explicitly regularizing the model in the weight space. The method is graphically illustrated in \cref{fig:framework}, and its algorithm is reported in \cref{alg:ours}. 
The key difference between our method to other model merging methods \citep{marouf2023weighted,kozal2024continual} is that our algorithm selectively merges the parameters that have shown large changes during training, which we assume that forgetting has occurred. Please note that this naive measure is possible owing to our unique setting (i.e., single source domain). Specifically, we selectively merge $80\%$ of the parameters that have shifted the most during training (compared to the snapshot $\theta_s$). The merging percentage was found using a grid search (see \cref{tab:selective}).

\section{Experiment}
\label{sec:exp}
\subsection{Experimental Setting}

\begin{table*}[t!]
\caption{Target domain accuracy on PACS and Digits.}
\label{tab:all_sdg_conventional}
\centering
\begin{adjustbox}{width=0.65\textwidth}
\centering\begin{tabular}{@{}lccccccccc@{}}
\toprule
& \multicolumn{4}{c}{{PACS}} & \multicolumn{5}{c}{{Digits}}\\
\cmidrule(lr){2-5} \cmidrule(lr){6-10}
{Method} & A & C & S & Avg. & SVHN & M-M & S-D & USPS & Avg.\\
\midrule
ERM & 54.43 & 42.74 & 42.02 & 46.39 & 27.83 & 52.72 & 39.65 & 76.94 & 49.29\\
ADA [\citenum{fan2021}] & 58.72& 45.58& 48.26& 50.85 & 35.51 & 60.41 & 45.32 & 77.26 & 54.62 \\
M-ADA [\citenum{qiao2020}] & \textbf{58.96} & 44.09 & 49.96 & 51.00 & 42.55 & 67.94 & 48.95 & 78.53 & 59.49\\
L2D [\citenum{wang2021}] & 56.26 & \textbf{51.04} & 58.42 & 55.24 & 62.86 & 87.30 & 63.72 & 83.97 & 74.46\\
PDEN [\citenum{li2021}] & 57.41 & 45.77 & 65.01 &  56.06 & 62.21 & 82.20 & 69.39 & 85.26 & 74.77\\
MetaCNN [\citenum{wan2022}] & 54.05 & 53.58 & 63.88 & 57.17 & \textbf{66.50} & \textbf{88.27} & 70.66 & 89.64 & 78.76\\
\midrule
RandAug. [\citenum{cubuk2020randaugment}] & 49.05 & 41.77 & 57.52 & 49.45 & 57.78 & 77.09 & 65.42 & 87.05 & 71.84\\
RandAug. + Mem. & 52.22  & 42.37 & 62.11 & 52.23 & 58.44 & 77.03 & 69.92 & 89.27 & 73.67\\
Ours & 56.95 & 47.99 & \textbf{70.20} & \textbf{58.38} & 63.93 & 77.79 & \textbf{80.54} & \textbf{93.42} & \textbf{78.92}\\

\bottomrule
\end{tabular}

\end{adjustbox}
\end{table*}

\begin{table*}[t!]
\caption{Target domain accuracy on larger datasets.}
\label{tab:all_sdg_new}
\centering
\begin{adjustbox}{width=0.77\textwidth} 
\centering\begin{tabular}{@{}lcccccccccccc@{}}
\toprule
& \multicolumn{4}{c}{{Office-Home}} & \multicolumn{4}{c}{{VLCS}} &\multicolumn{4}{c}{{Terra Incognita}}\\
\cmidrule(lr){2-5} \cmidrule(lr){6-9} \cmidrule(lr){10-13}
{Method} & Art & Clipart & Product & Avg. & L & C & S & Avg. & L38 & L43 & L46 & Avg.\\
\midrule

ERM & 52.78 & 40.19 & 68.73 & 53.90 & 59.06 & 97.30 & \textbf{74.25} & 76.87 & 22.90 & \textbf{15.85} & 22.91 & 20.55 \\
      
L2D$^\ast$ \citenum{wang2021} & 54.02 & 41.77 & 66.30 & 54.03 & 56.21 & 95.52 & 66.90 & 72.87 & 34.82 & 14.23 & 21.76 & 23.60 \\ 

PDEN$^\ast$ \citenum{li2021} & 53.39 & 43.38 & 66.25 & 54.34 & 62.55 & 96.11 & 73.52 & 77.39 & \textbf{37.52} & 14.93 & 20.80 & 24.42\\
      
\midrule
      
RandAug. $^\ast$ \citenum{cubuk2020randaugment} & 43.10 & 45.47 & 61.67 & 50.01 & 57.58 & 93.18 & 66.56 & 72.44 & 36.41 & 12.80 & 20.41 & 23.21\\

RandAug. + Mem. & 46.18 & 45.83 & 64.09 & 52.03 & 59.56 & 95.14 & 66.83 & 73.84 & 36.64 & 12.61 & 22.75& 24.00 \\
      
Ours$^\ast$ & \textbf{55.94} & \textbf{53.10} & \textbf{69.33} & \textbf{59.46} & \textbf{65.82} & \textbf{98.01} & 70.69 & \textbf{78.17} & 37.45 & 15.29 & \textbf{25.63} & \textbf{26.12} \\
\bottomrule
\end{tabular}

\end{adjustbox}
\end{table*}

\paragraph{Dataset}
There are a number of single-source domain generalization benchmarks, many of them deriving from domain generalization benchmarks. 

\textbf{{PACS}} \citep{pacs} consists of four domains of varying styles (Photo, Art, Cartoon, and Sketch) with 7 classes. Following previous works, we train our model with the Photo domain and evaluate the remaining target domains.  We use the train/test split provided by the original paper \citep{pacs}. \textbf{Digits} is also a very common sDG benchmark, composed of 5 different digit classification datasets, MNIST \citep{deng2012mnist}, SVHN \citep{svhn}, MNIST-M \citep{mnistm}, SYNDIGIT \citep{syndigit}, USPS \citep{usps}. In our experiment, we train our model with the first 10,000 samples of the MNIST dataset and assess the model's average accuracy across the remaining four domains. 

We also include larger benchmarks e.g., Office-Home \citep{officehome} VLCS \citep{fang2013unbiased} and Terra Incognita \citep{beery2018recognitionterraincognita}.
\textbf{Office-Home} is a common multi-DG benchmark consisting of 4 datasets (Real-world, Art, Clipart, Product) with differing styles with 65 classes. We train on the Real-world domain and evaluate the remaining domains. 
\textbf{VLCS} is also a benchmark for multi-DG, comprised of 4 datasets, PASCAL-VOC (V), LabelMe (L), Caltech-101 (C), and SUN09 (S) with varying styles. We used the PASCAL-VOC dataset as the source and the rest as target domains. 
\textbf{Terra Incognita} is comprised of 4 datasets of wildlife photos taken in different locations (L38, L43, L46, L46). We used the L38 dataset as the source and the rest as targets. 

\begin{figure*}[ht!]
    \centering
    \subfloat[Digits]{
        \includegraphics[width=0.48\textwidth]{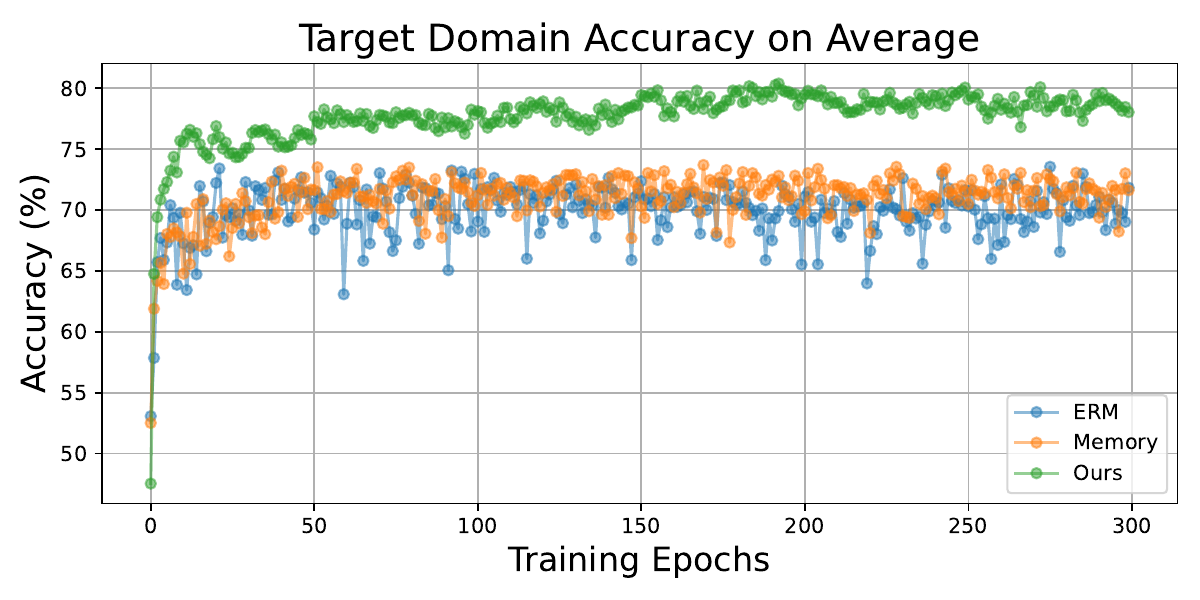}
        \label{subfig:digits_avg}
    }
    \hfill
    \subfloat[PACS]{
        \includegraphics[width=0.48\textwidth]{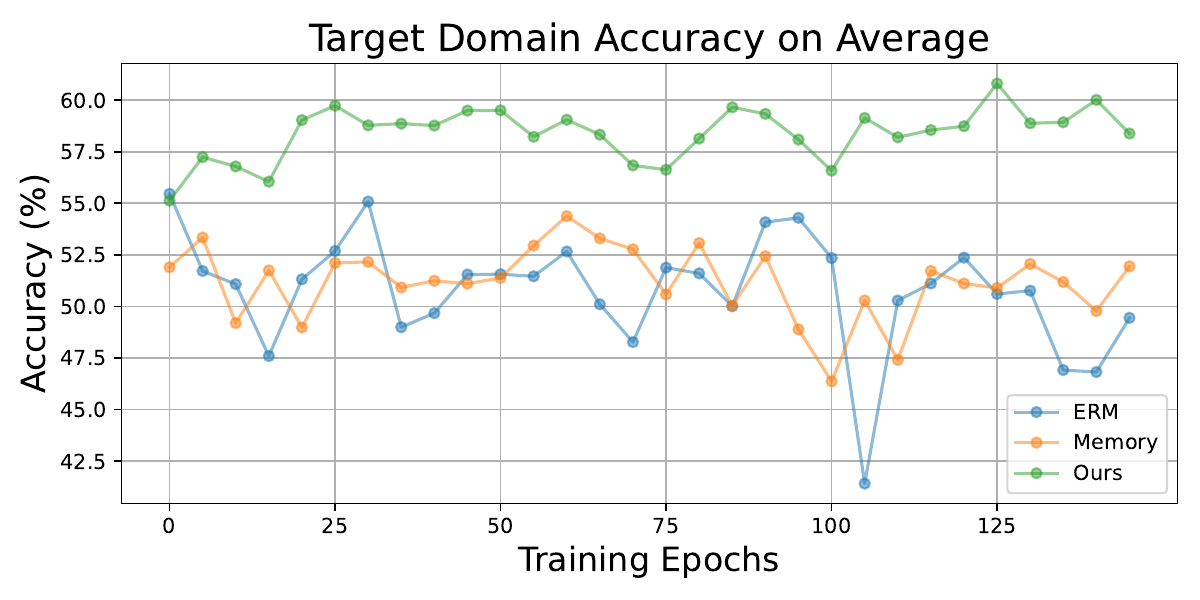}
        \label{subfig:pacs_avg}
    }
    \caption{Average target domain accuracy on the Digits (left) and PACS (right) datasets.}
    \label{fig:combined_avg}
\end{figure*}

\begin{table*}
  \centering
  \caption{Effect of selective merging on target domain accuracy}
  \label{tab:selective}
  \begin{adjustbox}{width=0.52\textwidth}
    \centering
\begin{tabular}{l|ccccc}
\toprule
\multicolumn{1}{l|}{} & \multicolumn{5}{c}{Datasets} \\
\cmidrule(l){2-6}

Merge $\%$ & PACS & Digits & Office-Home & VLCS & Terra Incognita  \\
\midrule

Full Merge & 77.82 & 57.01 & 59.23 & 76.95 &  24.47\\
Top $20\%$ & 75.11 & 56.48 & 57.94& 75.84 & 24.71\\
Top $40\%$ & 76.44& 57.45 & 59.80 & 76.82 & 24.32\\
Top $60\%$ & 77.69& 58.11 & 58.31 & 77.85 & 26.48\\
Ours (top $80\%$) & 78.92 & 58.38 & 59.46 & 78.17 &  26.12\\

\bottomrule
\end{tabular}

  \end{adjustbox}
\end{table*}

\begin{figure*}[ht!]
     \centering
     \hfill     
     \begin{subfigure}[b]{0.32\textwidth}
         \centering
         \includegraphics[clip,trim=1.6cm 3mm 3.6cm 1cm, height=3cm]{images/cka_erm_120.pdf}
         \caption{Without Memory.}
     \end{subfigure}
     \hfill
     \begin{subfigure}[b]{0.32\textwidth}
         \centering
         \includegraphics[clip,trim=1.6cm 3mm 3.6cm 1cm, height=3cm]{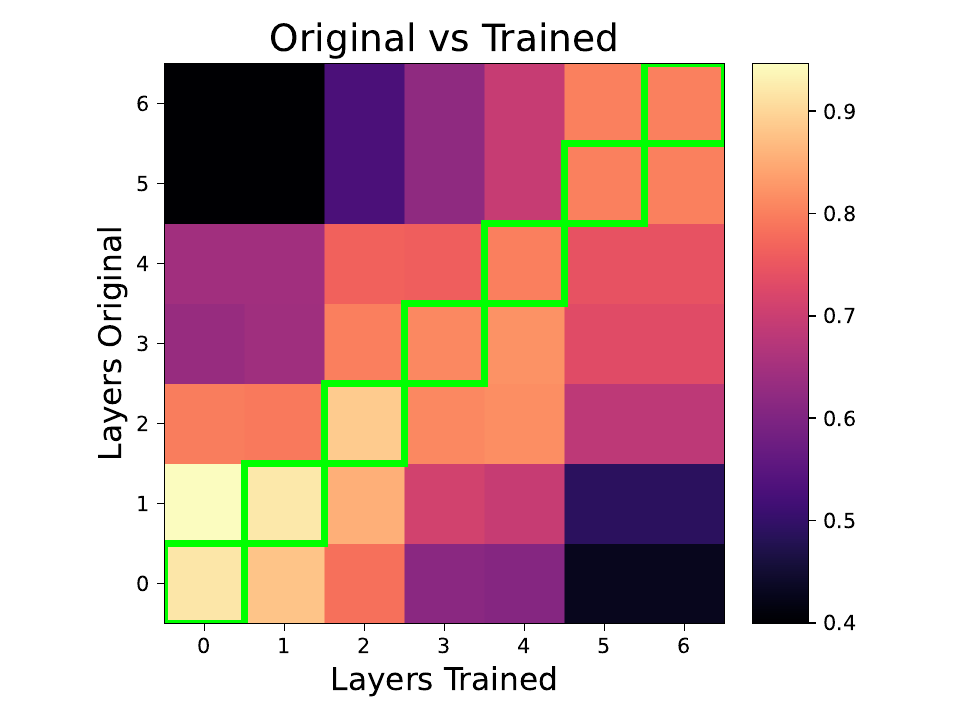}
         \caption{With Memory.}
     \end{subfigure}
    \hfill
     \begin{subfigure}[b]{0.32\textwidth}
         \centering
         \includegraphics[clip,trim=1.6cm 3mm 1.6cm 1cm, height=3cm]{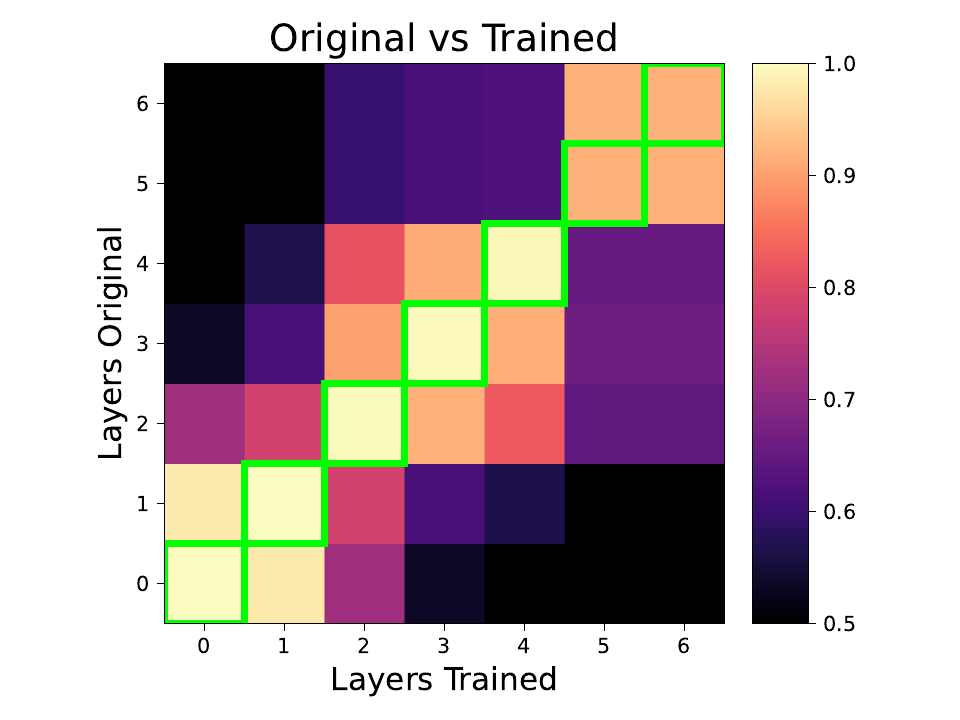}
         \caption{With Ours.}
     \end{subfigure}
     \hfill\null
        \caption{Layer-wise Feature Similarity (CKA) of the model between two models trained with two distinct augmentations of the same dataset. (1) Vanilla training (2) Training with replay memory (3) Training with our method. Brighter diagonal boxes indicate the robustness against forgetting.}
        \label{fig:cka_erm_memory}
\end{figure*}    

\paragraph{Model Architecture} 

Following previous works on sDG, we match the model architecture for a fair comparison with existing methods \citep{qiao2020,li2020domain,wan2022}. Specifically, for PACS, we use an AlexNet \citep{krizhevsky2012imagenet} architecture, while for Digits, we use a $3$-layer MLP. For larger benchmarks, we used the ResNet-(18/50) \citep{he2016deep} architecture.

\paragraph{Evaluation metrics}
Following previous sDG works, we evaluate the model's average accuracy across multiple target domain datasets. We will also report the accuracy of each target domain. Please refer to  \cref{tab:all_sdg_conventional} for a better understanding of the evaluation metric.


\subsection{Main Results}

We share our main experimental results in \cref{tab:all_sdg_conventional}. In PACS and Digits, our method with Random Augmentation displayed strong average accuracy, outperforming many previous works that utilize complex, learnable augmentation modules. More importantly, it is worth noting that a simple model merging procedure can significantly boost the generalization effect of random augmentation ($49.45 \rightarrow 58.38$ in PACS, $71.84 \rightarrow 78.03$). Furthermore, we share the experimental results on larger benchmarks in \cref{tab:all_sdg_new}. Please note that the Office-Home, VLCS, and Terra Incognita are not conventional benchmarks in the sDG literature, hence we could only compare with methods that we could replicate. Nevertheless, in all benchmarks, our selective merging approach showed effectiveness in boosting the generalization effect of random augmentation, showing competitiveness against all competing methods.

Furthermore, we report the results of the ablation study of our selective merging process. Specifically, we perform a study on the percentage of how many parameters are merged. The results of this experiment are shared in \cref{tab:selective}. We observe that the selective merging of the top $60-80\%$ parameter (ranked by the activeness of the parameter) outperforms others. However, there is a lack of understanding of why such a simple metric can be used as a selection criterion for merging. We believe further study is required on how to design reliable criteria for selective merging. 

Lastly, we show the effect of continual learning algorithms (e.g., replay memory, ours) on preserving the trained features. Similar to \cref{fig:cka_augmentation}, we measure the layer-wise feature similarity between two models (a) a model trained with a set of random augmentations $T_{1}$, and (b) the first model retrained on a set of non-identical set of augmentations $T_{2}$. Using this, we can show the model's robustness against forgetting induced by random augmentation. Specifically, we compare three cases (1) vanilla training without replay memory, (2) training with replay memory, and (3) training with our merging method. In \cref{fig:cka_erm_memory}, we show that our method is significantly better in addressing the forgetting (or feature distortion).



\section{Limitations \& Future Work}
In this section, we discuss the limitations of our work and propose future directions for our research. We begin this discussion by mentioning some remaining limitations. First, our theoretical framework currently assumes a finite, discrete set of augmentation transformations (\cref{appendix:augmentation_diversity}). This assumption may restrict the full potential of augmentation diversity. Extending our framework to allow more expressive perturbations—such as perturbing latent representations directly—could provide a richer variety of augmented samples, potentially improving out-of-distribution robustness further. Additionally, our experiments have primarily focused on classification benchmarks with limited domain diversity. Future work could examine more realistic or large-scale datasets, as well as non-classification tasks (e.g., detection or segmentation), to evaluate whether the proposed solution generalizes to more complex settings. Exploring these directions could yield deeper insights into how random data augmentation can be harnessed for more robust and versatile model training. Lastly, our partial merging method relies on some degree of heuristics that require manual adjustments (e.g. the measure of the parameter drift \cref{eq:drift}, the selection of the top $p\%$ parameters to be reset). We can think of an adaptive approach that would improve the method's robustness. We aim to reflect these ideas in our final copy.


\section{Concluding Remarks}
In this paper, we investigate why random augmentation underperforms in enhancing generalization in a limited data setting. Our findings reveal that the stochastic nature of random augmentation induces significant catastrophic forgetting during training. While continual learning strategies can mitigate this issue, their naive application in single-source domain generalization has limitations. To address this, we proposed a simple model merging method that counteracts this phenomenon, effectively leveraging random augmentation for improved generalization. Our approach achieved strong performance across multiple benchmarks—PACS, Digits, VLCS, Office-Home, and Terra Incognita-while significantly reducing computational costs.

\newpage
{
    \small
    \bibliographystyle{ieeenat_fullname}
    \bibliography{main}
}

\appendix

\clearpage

\section{Quantifying Augmentation Diversity: Random vs. Targeted Augmentation}\label{appendix:augmentation_diversity}

\paragraph{Entropy of Random Augmentation and Targeted Augmentation}

For a random augmentation $\tau_{ra}$, each transformation is selected with uniform probability $\frac{1}{n}$. Therefore, the entropy (augmentation diversity) in this case is: 
\begin{equation}\label{eq:ra_uniform}
    H(\tau_{ra}) = - \sum_{i=1}^{n} \frac{1}{n}\log \Bigl(\frac{1}{n}\Bigl) = \log n.
\end{equation}    
This entropy is maximized since all transformations in $T$ are equally probable under $\tau_{ra}$, resulting in the highest possible diversity in transformation selection. 

For targeted augmentation $\tau_{l}$, transformations are selected based on their loss values $\mathcal{L}(\theta; T_i(x))$, resulting in a non-uniform probability distribution. The entropy here is: 
\begin{equation} 
    H(\tau_{l}) = - \sum_{i=1}^{n} P_{\tau_{l}}(T_i)\log P_{\tau_{l}}(T_i).
\end{equation}    
As $P_{\tau_{l}}(T_i)$ assigns higher probabilities to transformations that optimize $\mathcal{L}$, the distribution is skewed, leading to $H(\tau_{l}) < H(\tau_{ra})=\log n$. The exact value of $H(\tau_{l})$ depends on the loss values and the hyperparameter $\beta$. As $\beta$ increases, the distribution becomes sharper, further reducing $H(\tau_{l})$.

\paragraph{Conclusion}

Since $H(\tau_{l}) < H(\tau_{ra})=\log n$, the entropy of the targeted augmentation is lower than that of the random augmentation. This indicates that random augmentation $\tau_{ra}$ provides greater diversity by uniformly exploring the full transformation set $T$, whereas targeted augmentation $\tau_{l}$ focuses on a subset of transformations that align with the objective $\mathcal{L}$. This reduced diversity in $\tau_{l}$ can lead to less exploration during training, potentially resulting in better optimization to $\mathcal{L}$ but possibly at the cost of the generalization effect. Conversely, the higher diversity in $\tau_{ra}$ may enhance generalization by exposing the model to a wider range of augmented samples, while it could also exacerbate unexpected issues, namely catastrophic forgetting.

\section{The Impact of Augmentation Diversity on Catastrophic Forgetting}\label{appendix:gradient_collision}

\paragraph{Recap on Augmentation Diversity}
In \cref{sec:limitation} and \cref{appendix:augmentation_diversity}, we showed that the random augmentation $\tau_{ra}$ uniformly selects the transformations from all $n$ possible transformations, resulting in diverse augmented samples $T(x)$. Consequently, the gradients $\nabla_{\theta}\mathcal(\theta;T(x))$ have high variance due to the diverse $T(x)$. On the other hand, the targeted augmentation $\tau_{l}$ selects its transformations based on the loss $\mathcal{L}$, leading to augmented samples $T(x)$ that are limited and closer to the original sample $x$ e.g., label-preserving augmentations \citep{geiping2022much}. This results in a lower variance in the gradients (i.e. gradient alignment).

\paragraph{Connecting Augmentation Diversity to Forgetting: Gradient Collision}

Now, we investigate how increased augmentation diversity can lead to a higher chance of catastrophic forgetting by focusing on gradient collision during training.

When training a model with parameters $\theta$ on augmented data, the parameter update is given by: $    \Delta\theta = - \eta \nabla_{\theta}\mathcal{L}(\theta;T(x)),$
where $\eta$ is the learning rate, and $T(x)$ the transformed version of $x$ using the augmentation set $T$. Let $g_x = \nabla_{\theta}\mathcal{L}(\theta;x)$ denote the gradient with respect to $x$, and $g_{T(x)} = \nabla_{\theta}\mathcal{L}(\theta;T(x))$ denote the gradient with respect to the augmented data $T(x)$.



To measure the alignment of the gradients $g_x$ and $g_{T(x)}$, we use the cosine similarity, formulated as: $    \cos(\phi) = \frac{g_x^{\intercal} g_{T(x)}}{\Vert g_x \Vert \cdot \Vert g_{T(x)} \Vert}, $
where $\cos(\phi) > 0$ indicates aligned gradients, and $\cos(\phi) < 0$ indicates gradient collision, which can contribute to catastrophic forgetting \citep{lopez2017gradient}.

Assuming the transformation $T(x)$ introduces a perturbation $\delta$ on the input $x$, we can write:
\begin{equation}
    T(x) = x + \delta,
\end{equation}

where $\delta$ is a random variable with zero mean ($\mathbb{E}[\delta]=0$) and covariance $\sum\nolimits_{T}$. Here, the covariance matrix $\sum\nolimits_{T}$ describes how the perturbations vary across different dimensions of the input space, its diagonal elements representing the variance of $\delta$ in each input dimension $i$.

Using a first-order Taylor expansion around $x$, we approximate $g_{T(x)}$ as:

\begin{equation}\label{eq:taylor}
    g_{T(x)} = \nabla_{\theta}\mathcal{L}(\theta; x + \delta)\approx \nabla_{\theta}\mathcal(\theta;x) + \nabla_{\theta}\nabla_{x}\mathcal{L}(\theta;x)\delta = g_x + M\delta,
\end{equation}
where $M = \nabla_{\theta}\nabla_{x}\mathcal{L}(\theta;x)$ is a matrix representing the mixed second-order derivatives.


Grounding on this, we analyze the expected cosine similarity (alignment) over $\delta$. We derive this by plugging in \cref{eq:taylor} to the cosine similarity $cos(\phi)$:

\begin{equation}\label{eq:expected_cosine_taylor}
    \mathbb{E}_{\delta}[\cos(\phi)] = \mathbb{E}_{\delta} \biggr[ \frac{g_x^{\intercal} (g_x + M\delta)}{\Vert g_x \Vert \cdot \Vert (g_x + M\delta) \Vert} \biggr].
\end{equation}

While the exact computation is complex, the trend of the expected cosine similarity can be analyzed by expanding the numerator and the denominator. 

\paragraph{Numerator of \cref{eq:expected_cosine_taylor}}

\begin{equation}
    g_x^{\intercal} (g_x + M\delta) = \Vert g_x \Vert^2  + g_x^{\intercal} M\delta.
\end{equation}

Since the expected value $\mathbb{E}_{\delta}[ g_x^{\intercal} M\delta ] = g_x^{\intercal}M\mathbb{E}[\delta]=0$ is zero, the expected numerator is $\Vert g_x \Vert^2$. The variance of $g_x^{\intercal}M\delta$ is:

\begin{equation}
    Var[g_x^{\intercal}M\delta] = g_x^{\intercal}M\sum\nolimits_{T}M^{\intercal}g_x.
\end{equation}

\paragraph{Denominator of \cref{eq:expected_cosine_taylor}}
In $\Vert g_x \Vert \cdot \Vert (g_x + M\delta) \Vert$,
\begin{equation}
    \Vert (g_x + M\delta) \Vert = \sqrt{ \Vert g_x \Vert^2 + 2g_x^\intercal M \delta + \Vert M\delta\Vert^2}.
\end{equation}

The expected value of $\Vert (g_x + M\delta) \Vert$ increases with the variance introduced by $\delta$, specifically due to $\Vert M\delta\Vert^2$. Consider a case where the transformation covariance $\sum\nolimits_{T}$ of $\delta$ increases:

\begin{itemize}
    \item Numerator: The variance of $g_x^{\intercal}M\delta$ increases, causing greater fluctuations around the expected value $\Vert g_x \Vert^2$.
    \item Denominator: The term $\Vert M\delta \Vert^2$ increases, leading to a larger denominator.
\end{itemize}

Overall, the expected cosine similarity $\mathbb{E}_{\delta}[\cos(\phi)]$  (\cref{eq:expected_cosine_taylor}) decreases because the denominator grows faster than the numerator. This decrease indicates increased gradient collision.

\paragraph{Conclusion}

An increase in the variance $\sum\nolimits_T$ of the transformation perturbations $\delta$ leads to decreased expected cosine similarity between $g_x$ and $g_{T(x)}$. This misalignment of gradients can result in gradient collision, effectively increasing the risk of catastrophic forgetting during parameter updates \citep{saha2023continual}. Therefore, higher augmentation diversity of random augmentation, characterized by larger perturbations, may exacerbate forgetting, whereas targeted augmentation with lower diversity suffers less from this issue.


\section{Why is model merging effective for continual learning in a single domain?}\label{appendix:analysis}

The continual learning effect of model merging (i.e. weight-averaging) has been discussed in the recent continual learning literature \citep{saha2023continual, kozal2024continual}. In general, combining the weights of neural networks trained on different tasks enables the model to retain knowledge from previous tasks while learning new ones. 

Weight-averaging helps in creating a parameter space that balances the representations learned from different tasks. By averaging the weights, the model finds a middle ground in the parameter landscape that is effective for multiple tasks \citep{izmailov2018averaging}. This results in a smoother loss surface, which enhances the model's generalization capabilities. Furthermore, model merging is analogous to ensemble methods, where combining multiple models leads to improved performance and robustness \citep{rame2022diverse}. In the context of continual learning, merging models trained on different tasks can be seen as creating an ensemble that captures diverse features from all tasks. 

Moreover, we view that weight averaging acts as an implicit regularizer, preventing the model from becoming too specialized on any particular augmented version of the data \citep{wen2016learningstructuredsparsitydeep}. This regularization is crucial for improving the model's performance on unseen data and ensuring stability during training. In summary, while augmented data does not constitute different tasks, model merging through weight averaging remains effective for continual learning within the same task. It facilitates the integration of knowledge from various data augmentations, enhancing the model's generalization capabilities and robustness.

\end{document}